\documentclass[12pt]{article}

\usepackage{sbc-template}

\usepackage[utf8]{inputenc}
\usepackage[brazil]{babel}
\usepackage{textgreek}
\usepackage{graphicx}
\usepackage{xpatch}
\usepackage{booktabs}
\usepackage[hyphens]{url}
\usepackage{amsmath}
\usepackage{hyperref}
\usepackage[table]{xcolor}
\definecolor{lightgreen}{RGB}{230, 245, 233}
\usepackage{array}

\usepackage{enumitem}

\newcolumntype{L}[1]{>{\raggedright\arraybackslash}p{#1}}

\title{Na Prática, qual IA Entende o Direito? Um Estudo Experimental com IAs Generalistas e uma IA Jurídica}

\author{Marina Soares Marinho\inst{1,2}, Daniela Vianna\inst{1}, Livy Real\inst{1,2},\\
Altigran da Silva\inst{2}, Gabriela Migliorini\inst{1} }

\address{Jusbrasil, Salvador - Brasil\\
  \vspace{-0.85cm}
\email{\{marina.marinho,gabriela.milgiorini,daniela.vianna\}@jusbrasil.com.br}
\vspace{-0.25cm}
\nextinstitute
  Instituto de Computação, Universidade Federal do Amazonas,
  Manaus - Brasil\\
   \vspace{-0.85cm}
  \email{\{{livyreal,alti\}@icomp.ufam.edu.br}}
}

\begin{document} 
\maketitle

\begin{abstract}
O uso de sistemas de inteligência artificial (IA) generativa no campo jurídico tem se expandido com rapidez, acompanhando o potencial de maior eficiência na pesquisa e na produção de textos jurídicos. Contudo, em grande parte dos casos, não é uma prática comum avaliar a qualidade do conteúdo jurídico apresentado nas respostas geradas por estes sistemas. Isso é fundamental, pois para que os usuários possam adotar estes sistemas, é importante aferir sua correção, consistência e adequação às formas próprias de construção e justificação do Direito. 
Visando a abordar este problema, este trabalho apresenta o \textit{Estudo Jusbrasil sobre o uso de IAs generalistas no Direito}, uma avaliação experimental comparativa entre três sistemas de IA aplicados ao Direito: Jus IA, ChatGPT e Gemini, examinando sua efetividade em tarefas projetadas para reproduzir o cotidiano profissional de advogados no Brasil. As respostas fornecidas por cada sistema foram avaliadas às cegas por 48 profissionais, com base em quatro critérios: corretude, completude, fluidez e confiabilidade.
Dentre os sistemas estudados, o Jus IA apresentou desempenho mais estável e superior em todos os critérios, com destaque para corretude e confiabilidade. Já os sistemas generalistas, desenvolvidos para atuar em múltiplos domínios, sem especialização jurídica, mostraram maior variação entre as métricas utilizadas e incidência de erros conceituais ou citações normativas inexistentes. O protocolo experimental adotado foi fundamentado em princípios da teoria do Direito, como correção material, coerência sistemática e integridade argumentativa, e em práticas profissionais jurídicas, assegurando neutralidade, reprodutibilidade e relevância analítica.
Estes resultados oferecem evidências de que a especialização no domínio é determinante para a qualidade e a confiabilidade das respostas construídas usando IA generativa.
Além disso, O estudo contribui para o avanço de uma abordagem metodológica que articula fundamentos dogmáticos e critérios empíricos na avaliação de sistemas de IA voltados à atuação jurídica.
\end{abstract}

\maketitle

\section{Introdução}

O uso de Modelos de Linguagem de Larga Escala (LLMs) em tarefas jurídicas tem se expandido rapidamente. Ferramentas baseadas em Inteligência Artificial (IA) generativa já são empregadas por escritórios, departamentos jurídicos e órgãos públicos para gerar minutas, revisar textos legais, sintetizar jurisprudência e responder consultas. Essa ampliação de uso evidencia o potencial transformador da tecnologia, ao mesmo tempo em que torna mais visíveis suas limitações quanto à validade e à confiabilidade das respostas jurídicas que produz. Neste sentido, \cite{hallufree}, ao avaliar plataformas comerciais de pesquisa jurídica baseadas em IA, constataram taxas significativas de respostas imprecisas e não verificáveis, mesmo em produtos de mercado amplamente difundidos. Sua pesquisa evidencia não apenas o desafio de avaliar conteúdo jurídico, mas a ausência de transparência metodológica e falta de padrões auditáveis de avaliação. 

A avaliação de qualidade é parte fundamental do desenvolvimento de produtos do Jusbrasil. A empresa é a maior \emph{legal tech} da América Latina e possui o maior repositório de documentos jurídicos do Brasil. Em março de 2025, a empresa lançou o Jus IA, um sistema de Inteligência Artificial generativa que, atualmente, funciona em modo conversacional e permite às pessoas usuárias interagirem para realizar, virtualmente, qualquer tarefa relacionada com o consumo ou criação de conteúdo jurídico. 

O raciocínio jurídico distingue-se de outros domínios especializados porque a validade de uma resposta depende não apenas da exatidão informacional, mas da capacidade de reconstruir o sentido das normas dentro de um sistema interpretativo e argumentativo que confere autoridade ao resultado. Assim, a qualidade de uma resposta jurídica não pode ser medida apenas pela similaridade textual ou pela verificação factual. O desafio é avaliar se a resposta é juridicamente correta, ou seja, se ela é coerente com o ordenamento e argumentativamente íntegra.

Hoje, pela novidade da tecnologia, faltam protocolos sistemáticos e transparentes que traduzam os fundamentos da racionalidade jurídica em critérios empíricos de avaliação de IAs. Como também observam \cite{hallufree}, mesmo as principais ferramentas jurídicas comerciais não divulgam metodologias de avaliação, o que impede auditoria científica e dificulta a criação de parâmetros públicos de comparação. Nos últimos anos, um conjunto relevante de estudos vem tentando aproximar lógica jurídica e métodos avaliativos, mas de forma ainda fragmentada e desarticulada. Pesquisas como as de \cite{xu2023qa} e \cite{mullick2022evaluation} propõem métricas inspiradas em estruturas argumentativas clássicas, como o modelo Issue–Reason–Conclusion e a extração de \textit{intents} jurídicos; \cite{posner2025judgeai} analisam a aderência dos LLMs a diferentes correntes de pensamento jurídico, revelando vieses de formalismo e realismo; e benchmarks como JuDGE~\cite{su2025judge}, CaseGen~\cite{li2025casegen} e LRAGE~\cite{park2025} experimentam formas de mensurar coerência, factualidade e precisão normativa por meio de avaliadores automáticos. No entanto, esses esforços permanecem dispersos, restritos a contextos específicos e sem um referencial comparável de qualidade entre sistemas jurídicos, línguas ou jurisdições.

Um destaque latino-americano nesse panorama é o OAB-Bench, proposto por \cite{pires2025oabbench}, que utiliza a segunda fase do Exame da Ordem dos Advogados do Brasil como benchmark de escrita jurídica aberta. O estudo avalia se LLMs seriam capazes de redigir petições e respostas jurídicas segundo os critérios oficiais da OAB, como correção doutrinária, estrutura argumentativa e aderência às normas processuais, e emprega um modelo de IA como juiz (\textit{LLM-as-a-Judge}) para atribuir notas de forma automatizada. A pesquisa demonstra que é possível traduzir critérios jurídicos normativos em métricas mensuráveis, mas também evidencia limitações na consistência dos avaliadores automáticos e na padronização de rubricas, reforçando a necessidade de protocolos mais transparentes e auditáveis para o domínio jurídico.
Além disso, como mostra o levantamento de práticas industriais \cite{rosenoff2025,heller2024,dahnHerrmannova2025}, as grandes lawtechs globais, como Harvey, Thomson Reuters e LexisNexis, vêm desenvolvendo seus próprios protocolos internos de avaliação, com métricas híbridas que combinam avaliadores automáticos (synthetic raters) e revisão humana. Embora avancem em escalabilidade e controle de riscos, esses processos não são públicos, auditáveis ou interoperáveis com os critérios da pesquisa científica, o que reforça a assimetria entre as avaliações corporativas e os padrões de transparência exigidos para governança pública e reprodutibilidade acadêmica.

O \textit{Estudo Jusbrasil sobre o uso de IAs generalistas no Direito} surge, portanto, não para inaugurar o tema da avaliação jurídica de IAs, mas para propor sua sistematização. Seu objetivo é construir um referencial metodológico e experimental unificado, capaz de articular princípios da teoria do Direito com práticas empíricas de avaliação de IA, de modo que a qualidade de sistemas generativos jurídicos possa ser mensurada de forma transparente, comparável e auditável.

A hipótese central é que sistemas especializados em tarefas jurídicas, treinados ou calibrados com dados normativos, jurisprudenciais e doutrinários, curados e auditados por profissionais da área, produzem respostas de maior qualidade para o domínio, caracterizadas por confiabilidade, coerência e aderência ao raciocínio jurídico. Ao estruturar métricas ancoradas na racionalidade jurídica, o \textit{Estudo Jusbrasil sobre o uso de IAs generalistas no Direito} busca preencher a lacuna metodológica entre o saber dogmático e o saber empírico, criando as bases para uma governança pública e científica da IA jurídica.

O presente artigo tem, portanto, três objetivos principais:
\begin{enumerate}
    \item Apresentar a fundamentação teórica que inspira o protocolo Jus IA, traduzindo a estrutura do raciocínio jurídico em critérios avaliativos.
    \item Descrever o desenho experimental e o método de avaliação comparativa entre uma IA jurídica e IAs generalistas.
    \item Discutir os resultados empíricos obtidos, explorando o papel da especialização na qualidade jurídica das respostas. \end{enumerate}

\section{Fundamentação Teórica e Princípios Orientadores}
\label{section-fundamentacao}


Esta seção apresenta os fundamentos jurídicos que orientam o protocolo de avaliação proposto, estabelecendo um elo entre a teoria do Direito e os critérios utilizados para mensurar a qualidade das respostas geradas artificialmente. A partir de referenciais como a integridade da argumentação, a coerência do raciocínio jurídico e a confiabilidade das fontes normativas, são explicitados os princípios que sustentam o desenho metodológico. Esses princípios se traduzem em critérios operacionais, tais como corretude, confiabilidade, completude e fluidez, cuja escolha é justificada com base em sua capacidade de capturar dimensões essenciais da prática jurídica.

\subsection{O Direito como sistema interpretativo}

O Direito é um sistema de produção normativa e discursiva que opera inteiramente no campo da linguagem. Seu funcionamento depende de práticas interpretativas e argumentativas que transformam textos normativos em decisões aplicáveis. Desde \cite{hart1961concept} e \cite{luhmann1993recht}, compreende-se que o Direito se mantém não pela rigidez das regras, mas pela reprodução institucional de expectativas normativas, mediadas por processos de interpretação e justificação. Cada decisão jurídica atualiza e reconstrói o sentido das normas, mantendo a coerência do sistema.

Essa estrutura faz do raciocínio jurídico uma prática essencialmente interpretativa, guiada por coerência e integridade~\cite{dworkin1986laws}: as respostas jurídicas não são deduções formais, mas reconstruções argumentadas do ordenamento. Como observa \cite{gadamer1960wahrheit}, interpretar é fundir horizontes; integrar o texto às condições históricas e institucionais do intérprete. O conhecimento jurídico, portanto, é linguístico, contextual e institucionalmente situado.

Essas características não inviabilizam o uso de IA generativa no Direito. Ao contrário, definem as condições sob as quais este uso pode ser aplicado e avaliável. Sistemas de IA especializados podem reproduzir e apoiar práticas jurídicas se forem projetados para compreender o Direito como sistema interpretativo, reconhecendo a centralidade da argumentação, da coerência e da justificabilidade pública. Avaliar sua qualidade, portanto, exige questionar em que medida as IAs preservam esses modos de raciocínio.

\subsection{A IA jurídica e o circuito hermenêutico}

Quando sistemas de IA generativa produzem textos jurídicos, eles não apenas processam dados, mas participam da produção de sentido jurídico, sintetizando interpretações existentes e recombinando fundamentos. Isso significa que sua qualidade deve ser avaliada não apenas por critérios técnicos (precisão e consistência, por exemplo), mas por sua aderência à racionalidade jurídica, isto é, pela capacidade de formular respostas interpretativamente defensáveis e normativamente justificadas.

A linha de pesquisa conhecida como Artificial Intelligence and Law oferece fundamentos para essa abordagem. \cite{ashley2017artificial} discute que o raciocínio jurídico é sempre argumentativo e contextual; sua formalização computacional não elimina essa natureza, mas a torna explícita. \cite{bench-capon2003model} complementam ao propor que a coerência do Direito decorre da ponderação entre valores e fundamentos, e não da aplicação mecânica de regras. Essa literatura demonstra que é possível e desejável que modelos de IA reproduzam estruturas argumentativas e padrões de coerência típicos do raciocínio jurídico, desde que se considerem as camadas normativas e discursivas que o constituem.

\subsection{Do raciocínio jurídico à avaliação de qualidade}

O \textit{Estudo Jusbrasil sobre o uso de IAs generalistas no Direito} parte dessa compreensão: avaliar um sistema de IA jurídica significa medir como ele se alinha à estrutura interpretativa e argumentativa do Direito. Por isso, seu protocolo traduz quatro dimensões centrais do raciocínio jurídico em critérios empíricos observáveis, como descritos na Tabela~\ref{tab:criterios_jusia}.

\begin{table*}[h!]
\centering
\small
\caption{Dimensões jurídico-argumentativas e critérios do protocolo \textit{jus IA}}
\rowcolors{2}{lightgreen}{white}
\begin{tabular}{m{6cm} m{6cm} m{2cm}} 
\toprule
\textbf{Dimensão jurídico-argumentativa} & \textbf{Descrição} & \textbf{Critério} \\ 
\midrule
\textbf{Correção material} & Compatibilidade da resposta com o ordenamento e sua aplicação contextual. & Corretude \\
\textbf{Coerência argumentativa} & Consistência entre fundamentos, premissas e conclusões; transparência da justificativa. & Confiabilidade \\
\textbf{Abrangência interpretativa} & Consideração adequada dos elementos normativos e fáticos relevantes. & Completude \\
\textbf{Aderência comunicativa} & Clareza, precisão e adequação terminológica à prática profissional. & Fluidez \\ 
\bottomrule
\end{tabular}
\rowcolors{2}{}{}
\label{tab:criterios_jusia}
\vspace{-0.5em}
\end{table*}

Esses critérios expressam o modo de produção de sentido jurídico, permitindo avaliar empiricamente a capacidade de um sistema generativo reproduzir práticas legítimas de interpretação. Assim, a avaliação de qualidade deixa de ser um simples controle técnico e se torna um procedimento de verificação interpretativa, compatível com a natureza discursiva do Direito.

\subsection{A qualidade como vínculo entre técnica e legitimidade}

Nessa perspectiva, a qualidade jurídica não é apenas um atributo das respostas, mas o elo que conecta a operação técnica da IA à legitimidade institucional do Direito. Avaliar qualidade é verificar se o sistema mantém coerência, justificabilidade e rastreabilidade em suas respostas (dimensões que sustentam a confiança pública na prática jurídica).

O \textit{Índice de Qualidade Jus IA} propõe, assim, uma abordagem hermenêutica e mensurável: combina critérios normativos do raciocínio jurídico com métodos empíricos de avaliação, permitindo comparar soluções e orientar seu aprimoramento. Em vez de limitar o uso da IA no Direito, essa estrutura o torna viável e auditável, estabelecendo parâmetros de confiabilidade que preservam a integridade das práticas interpretativas.

\section{Metodologia}

Esta seção apresenta o desenho experimental e os procedimentos metodológicos adotados para a avaliação comparativa entre sistemas generativos no contexto jurídico. São detalhadas as cinco intenções de uso que orientam a construção dos casos, bem como as estratégias de controle de viés, anonimização e randomização das respostas. Em seguida, descreve-se a composição da amostra, incluindo o número de casos, tarefas jurídicas e sistemas comparados, e o perfil dos avaliadores envolvidos. Por fim, são explicitados os instrumentos de coleta e os métodos de análise estatística e qualitativa empregados na interpretação dos resultados.

\subsection{Desenho geral do experimento}
\label{sec-desenho-exp}

O experimento comparou quatro sistemas de IA conversacional baseados em LLMs:

\begin{enumerate}
    \item IA Jurídica (Jus IA): sistema especializado ajustado com corpora normativos, doutrinários e jurisprudenciais brasileiros, voltado a interpretação e argumentação jurídica contextualizada\footnote{Acessado via \url{ia.jusbrasil.com.br}}.
    \item ChatGPT: sistema de propósito geral com ampla difusão comercial\footnote{Acessado via \url{chat.openai.com}}. Foram avaliados duas versões: a versão gratuita (ChatGPT Gratuito) e a plus (ChatGPT Plus), todas com o modelo GPT-5. 
    \item Gemini: sistema de mesma geração tecnológica\footnote{Acessado via \url{gemini.google.com/app}}. Durante a avaliação do sistema foi selecionado o modelo 2.5 PRO. 
\end{enumerate}

A escolha das soluções comparadas partiu do reconhecimento de que ChatGPT e Gemini são as duas IAs generalistas mais utilizadas no mundo\footnote{\url{https://aimagazine.com/news/top-10-generative-ai-tools}} e têm também forte presença no Brasil. O país figura entre os que mais acessam o ChatGPT globalmente, com mais de 140 milhões de interações diárias, segundo levantamento da SimilarWeb divulgado pela CNN Brasil\footnote{\url{cnnbrasil.com.br/tecnologia/brasil-esta-entre-os-paises-que-mais-usam-o-chatgpt-diz-estudo/}}. O Gemini, lançado oficialmente no país em dezembro de 2023, vem ampliando o uso de modelos multimodais e se consolidando como alternativa relevante entre profissionais do Direito.
    
Cada sistema foi avaliado em cinco \textbf{\textit{intenções de uso}} representativas da rotina jurídica. No contexto do experimento, denomina-se \textit{intenção de uso} o propósito funcional da interação com a IA, isto é, o tipo de atividade profissional que motiva a interação com a ferramenta e determina o formato esperado da resposta. São elas:

\begin{enumerate}
\item \textbf{Gerar / Ajustar Documento}: produção ou revisão de trechos ou documentos completos tipicamente utilizados na advocacia (petição, notificação, contrato, parecer etc.).
\item \textbf{Analisar / Resumir Documento}: síntese e avaliação crítica de documentos jurídicos.
\item \textbf{Entender Conceito Jurídico}: explicação de categorias e institutos, com exigência de definição, enquadramento normativo e exemplos.
\item \textbf{Entender Panorama Jurisprudencial}: mapeamento de entendimentos predominantes, com identificação de \textit{leading cases} e eventuais divergências entre órgãos e tribunais.
\item \textbf{Pesquisar Precedente}: localização e referência a precedentes aplicáveis ao caso da pessoa usuária, com indicação do raciocínio quanto à pertinência dos documentos elencados.\end{enumerate}

Cada intenção foi avaliada com três casos jurídicos distintos, totalizando 15 casos. Para cada caso foram avaliadas quatro saídas, uma de cada sistema de IA: Jus IA, ChatGPT Gratuito, ChatGPT Plus e Gemini, totalizando 60 saídas a serem avaliadas.

\subsection{Construção dos casos jurídicos e das instruções do usuário}

Os casos jurídicos utilizados no experimento foram elaborados a partir de interações reais registradas na plataforma Jus IA, previamente tratadas em conformidade com a Lei Geral de Proteção de Dados (LGPD). Todo o material é integralmente anonimizado, de modo que nenhum dado pessoal, identificação de partes ou contexto específico das conversas seja acessível aos sistemas de IA ou aos avaliadores. 

Após o tratamento preliminar, as instruções transmitidas aos modelos foram revisadas por pessoas advogadas e pesquisadoras da equipe Jusbrasil, garantindo precisão técnica, neutralidade e representatividade dos casos avaliados.


Cada intenção de uso foi previamente associada a um conjunto fixo de três casos, selecionados para variar em complexidade e extensão das respostas esperadas, de modo a refletir a diversidade de tarefas que compõem a rotina profissional jurídica.

Todos os sistemas foram submetidos às mesmas instruções e aos mesmos materiais de caso. As instruções foram redigidas de forma idêntica e revisados para eliminar possíveis vieses de formulação, enquanto os materiais podem incluir documentos anexados específicos a uma tarefa.  Todas as ferramentas foram utilizadas em seu modo de interação conversacional, assegurando condições equivalentes de uso e foco em sistemas conversacionais (e não em aplicações para tarefas jurídicas específicas). Dessa forma, as diferenças observadas entre os sistemas refletem sua capacidade de compreender a intenção da tarefa, antecipar o formato esperado da resposta e estruturar adequadamente o conteúdo jurídico produzido.

O dataset desenvolvido como parte deste trabalho está disponível publicamente no Hugging Face\footnote{\url{https://huggingface.co/datasets/jusbrasil/indice-jusia}}, visando facilitar reprodutibilidade e reuso.

\subsection{Pesos e hierarquia dos critérios de avaliação}
\label{sec:pesos-criterios}

A definição e ponderação dos critérios de avaliação do \textit{Protocolo Jus IA} resultaram da combinação entre experiência profissional de pessoas advogadas envolvidas no projeto, do conhecimento acumulado em pesquisas de usuário conduzidas pela equipe do Jusbrasil ao longo dos últimos anos e também foi resultado de um formulário respondido pelos avaliadores, em que eles deveriam apontar quais são os critérios mais relevantes para determinar a qualidade de um conteúdo jurídico. O objetivo foi garantir que o índice expressasse não apenas fundamentos teóricos de correção e coerência, mas também as preocupações históricas e pragmáticas da advocacia brasileira no uso de sistemas de informação jurídica. Dessa síntese emergiu uma hierarquia empírica de relevância entre os critérios, refletindo o modo como profissionais do Direito avaliam a qualidade de um texto jurídico produzido por terceiros, humanos ou artificiais.

Os critérios propostos durante a avaliação foram: \textit{confiabilidade}, \textit{corretude}, \textit{completude} e \textit{fluidez}.
Os critérios de \textit{confiabilidade} e \textit{corretude} receberam peso interpretativo maior no processo de avaliação, por representarem o núcleo das expectativas profissionais quanto à precisão da informação jurídica e à rastreabilidade das fontes. Essas dimensões correspondem, no plano teórico, às exigências de consistência e integridade descritas na Seção~\ref{section-fundamentacao}: são os elementos que permitem que uma resposta seja reconhecida como juridicamente defensável e institucionalmente legítima. A ênfase nos critérios de confiabilidade e corretude aproxima-se dos parâmetros adotados por \cite{hallufree} et al. (2024) em sua análise das ferramentas de IA jurídica comerciais, centrada na rastreabilidade das fontes e na precisão normativa das respostas.

Na prática da advocacia, a confiabilidade traduz a necessidade de poder verificar e revisar o raciocínio exposto; a corretude assegura que o resultado esteja alinhado ao ordenamento vigente e à interpretação dominante. Ambos respondem, portanto, à função de legitimar o conteúdo jurídico como base para a ação profissional, o que explica seu peso superior no julgamento global de qualidade.

O critério de \textit{completude}, embora relevante, foi tratado com menor peso relativo. Em contextos conversacionais, a completude de uma resposta pode ser progredida por meio de interações subsequentes, à medida que o usuário refina ou amplia o escopo da pergunta. Além disso, a noção de completude envolve certo grau de subjetividade, sendo dependente das preferências individuais quanto ao nível de detalhe, extensão e granularidade desejados. Por isso, considerou-se mais adequado avaliá-la como um indicativo de abrangência contextual, e não como requisito absoluto de qualidade.

Por fim, a \textit{fluidez}, entendida como clareza, coesão e naturalidade linguística, foi incluída no protocolo como critério complementar, sem peso predominante.
Nos testes preliminares e nas observações de uso, verificou-se que as diferenças de fluidez entre sistemas são mínimas, pois todos compartilham modelos fundacionais de linguagem baseados em arquiteturas semelhantes (LLMs de última geração). Em outras palavras, a fluidez já constitui um atributo estrutural da tecnologia atualmente, pouco discriminativo entre ferramentas, e dificilmente compromete o valor jurídico de uma resposta. Assim, a função principal da fluidez na métrica é atuar como contrapeso comunicativo: assegurar que respostas juridicamente precisas também sejam legíveis e utilizáveis no contexto profissional.

A partir então dos quatro critérios predefinidos, obtemos a seguinte fórmula para o \textit{Índice Jus IA}:

\begin{small}
    \begin{equation}
\label{eq:desempenho}
\text{desempenho} = (2 \cdot \text{corretude}) + \text{ completude } + (2 \cdot \text{confiabilidade}) + \text{ fluidez}
\end{equation}
\end{small}

Essa distribuição de pesos reforça o princípio, discutido na fundamentação teórica, de que a qualidade jurídica é inseparável da legitimidade interpretativa. Avaliar sistemas de IA sob essa ótica significa reconhecer que a excelência técnica, expressa em fluência ou completude, só se torna relevante quando sustentada por confiabilidade e correção normativa.

A atribuição de pesos às respostas binárias (“sim”, “não” e “N/A”) variou conforme a intenção de uso, refletindo nuances metodológicas específicas de cada tarefa. Para o grupo “Gerar / Ajustar documento”, respostas afirmativas receberam peso 1, enquanto negativas e não aplicáveis foram penalizadas com peso -1, dado o caráter obrigatório da citação, com exceção do caso 3 (escrita de contrato), em que determinadas perguntas foram desconsideradas por não se aplicarem ao contexto. Por exemplo, citações não são necessária na criação de um contrato. 
Na intenção “Analisar e resumir documentos”, todas as respostas foram ponderadas de forma neutra (peso 1 para “sim” e 0 para “não” e “N/A”), já que a citação não era mandatória. Para as demais intenções, como “Entender conceito jurídico”, “Entender panorama jurisprudencial”, “Pesquisar precedentes/casos similares” e “Buscar documento exato”, foi adotado um esquema híbrido: perguntas relacionadas a citação (2, 3, 7 e 8) seguiram a lógica de penalização (-1 para “não” e “N/A”), enquanto a pergunta 9, relacionada a recência da citação, foi tratada com neutralidade (peso 0 para “não” e “N/A”), evitando dupla penalização em casos onde a citação não era obrigatória ou a pergunta era secundária. As perguntas são apresentadas no Apêndice~\ref{sec:apendiceA}.

\subsection{Controle experimental}

Para mitigar vieses, foram adotados os seguintes controles:
\begin{itemize}
    \item \textbf{Cegamento duplo adaptado:} os avaliadores não foram informados da origem dos conteúdos (IA jurídica ou generalista), nem que as respostas haviam sido geradas por ferramentas de IA generativa;
    \item \textbf{Randomização da ordem de exibição:} as respostas geradas pelos sistemas foram apresentadas aos avaliadores em sequência aleatória, de modo que a posição de cada resposta variava a cada caso;
    \item \textbf{Padronização de instruções:} todos os avaliadores receberam guias idênticos sobre o uso dos critérios e a forma de execução da tarefa;
    \item \textbf{Independência entre avaliadores:} cada avaliador realizou sua análise de forma independente, sem acesso às respostas atribuídas pelos demais participantes;
    \end{itemize}
    
Reitera-se que as respostas foram mantidas tal como geradas por cada sistema, sem intervenções sobre extensão ou formatação. Além de a própria interface conversacional não permitir ajustes desse tipo, características como o tamanho e a formatação do texto gerado são dos critérios que mais impactam a avaliação de modelos de IA~\cite{chen2024humans}. O procedimento de duplo cegamento, aliado à avaliação autônoma do critério de fluidez, contribuiu para mitigar possíveis vieses de preferência por determinado estilo ou formato de resposta. 

O desenho experimental segue boas práticas internacionais de avaliação empírica em IA jurídica, como as adotadas por \cite{hallufree}, que aplicam avaliação cega por múltiplos analistas e, quando necessária, uma terceira revisão independente para mitigar vieses de julgamento~\cite[Capítulo 13]{BPLN_livro_3ed:2024}. 

Esses procedimentos contribuíram para garantir comparabilidade e validade interna ao experimento, reduzindo a influência de fatores extrínsecos como familiaridade com linguagem ou estilo dos modelos.

\subsection{Perfil dos avaliadores}

Participaram do experimento 48 profissionais do Direito, provenientes de diferentes regiões do Brasil e com perfis complementares de formação e atuação. Essas pessoas foram selecionadas a partir de um banco de cadastro permanente, mantido pelo Jusbrasil. Originalmente, foram selecionadas 50 pessoas colaboradoras, no entanto, duas participações foram excluídas após a análise de concordância entre os avaliadores, por apresentarem divergências substanciais em relação às demais respostas.

A amostra abrangeu pessoas advogadas da iniciativa privada (62.5\%), servidores públicos (20.8\%), docentes e pesquisadores (10.4\%) e membros de carreiras jurídicas diversas (6.3\%), como magistratura e Ministério Público.  
O grupo apresentou tempo médio de experiência profissional de nove anos, com variação entre recém-formados e profissionais com mais de dez anos de prática. Essa diversidade foi intencionalmente buscada para representar diferentes perspectivas sobre a aplicação do Direito na prática cotidiana. 

Em relação às áreas de especialização, predominam o Direito Civil (27\%), o Direito do Trabalho (18\%) e o Direito Administrativo (15\%), seguidos por Direito Penal, Empresarial, Constitucional e Tributário, distribuídos de forma coerente ao perfil do público que utiliza ferramentas de apoio jurídico na Plataforma do Jusbrasil.  

Os avaliadores estão distribuídos em 12 unidades federativas, com concentração nas regiões Sudeste (46\%), Sul (23\%) e Nordeste (17\%), além de participações pontuais das regiões Centro-Oeste (10\%) e Norte (4\%). 
Essa dispersão geográfica reforça a heterogeneidade das interpretações e práticas jurídicas representadas na amostra.

Todos os participantes atuam regularmente em funções que envolvem interpretação e aplicação do Direito (redação de peças, análise de documentos, consultoria ou ensino) e foram previamente instruídos sobre os critérios do protocolo \textit{Jus IA}, participando de uma rodada piloto de calibragem antes das avaliações formais.  

\subsection{Instrumento de coleta}

O formulário de avaliação, implementado através da plataforma Retool\footnote{\url{https://retool.com}}, contém:
\begin{itemize}
    \item dez campos binários distribuídos entre os critérios de avaliação: três questões de corretude, quatro de confiabilidade, duas de completude e uma de fluidez;
    \item um campo qualitativo aberto para comentários;
    \item registro automático de tempo de análise e de ordem de apresentação.
\end{itemize}

Os campos binários receberam essa denominação porque restringiam a resposta às opções ``sim'', ``não'' e ``não se aplica''. O julgamento seguiu uma lógica de \textit{tudo ou nada}: qualquer ocorrência de incorreção, lacuna, inconsistência ou falha de fluidez ao longo da resposta resultava na marcação negativa do critério correspondente. Assim, apenas as respostas sem nenhuma ocorrência adversa eram consideradas corretas, completas, confiáveis e fluentemente redigidas.

A escolha pelo formato binário (sim/não) foi motivada por duas razões metodológicas:
\begin{itemize}
    \item forçar o avaliador a um julgamento conclusivo, evitando gradientes subjetivos;
    \item facilitar a agregação estatística e o cálculo de concordância interavaliador~\cite{BPLN_livro_3ed:2024}.\end{itemize}

\subsection{Procedimentos de análise}

As respostas foram analisadas em três níveis:
\begin{enumerate}
    \item Quantitativo-descritivo: cálculo de proporções positivas por critério, modelo e intenção de uso;
    \item Comparativo: diferença percentual média entre modelos;
    \item Qualitativo: análise temática das justificativas, com codificação em categorias emergentes como ``precisão normativa'', ``estrutura argumentativa'', ``clareza comunicativa'' e ``erro conceitual''.
\end{enumerate}


Também foi calculado o índice de concordância interavaliador, o coeficiente $\alpha$ de Krippendorff~\cite{Krippendorff}. O coeficiente de Krippendorff é uma métrica estatística usada para avaliar a confiabilidade entre avaliadores, especialmente em contextos onde múltiplas pessoas julgam ou classificam dados de forma subjetiva. Ele é particularmente útil em tarefas de anotação linguística, análise de conteúdo e avaliação de sistemas de IA. Ele quantifica a proporção de concordância observada que excede o que seria esperado por acaso, variando de 1.0 (concordância perfeita) a valores negativos (menos concordância que o acaso, indicando ruído ou ambiguidade). No presente estudo, com 48 avaliadores, foi obtida uma média de $alpha=0,6$, o que indica um nível de consistência substancial entre os julgamentos. Esse resultado reforça a confiabilidade do protocolo de avaliação adotado.






\section{Resultados e Discussão}

Nesta seção, apresentam-se os resultados da avaliação empírica dos sistemas, organizados por critério (corretude, completude, confiabilidade e fluidez) e por intenção de uso jurídico. A análise contempla tanto os dados quantitativos obtidos a partir dos campos binários quanto as variações observadas entre os modelos em diferentes contextos de aplicação, como geração de documentos, análise crítica, explicação de conceitos, mapeamento jurisprudencial e pesquisa de precedentes. Além dos indicadores estruturados, destacam-se observações qualitativas relevantes, especialmente comentários dos avaliadores jurídicos que ajudam a contextualizar nuances de estilo, clareza argumentativa e adequação normativa nas respostas geradas.

\subsection{Desempenho geral}
\label{sec-desenpenho-geral}

De forma geral, o Jus IA, sistema de IA especializado, apresentou o melhor desempenho entre os sistemas avaliados, com as maiores médias em todos os critérios: \textit{corretude} (0,84), \textit{completude} (0,63), \textit{confiabilidade} (0,83) e \textit{fluidez} (0,77). Além dos valores mais altos, destacou-se pela consistência entre os casos, com menor variação nos resultados. Esse comportamento é ilustrado na Tabela~\ref{tab:resultado_geral_criterios} que traz as médias e o desvio padrão dos sistemas de IA em cada um dos critérios de avaliação.

O desvio padrão indica o grau de dispersão dos resultados: quanto menor esse valor, mais próximos entre si estão os desempenhos observados. Assim, um sistema com baixo desvio padrão demonstra respostas mais uniformes e previsíveis, enquanto valores mais altos sugerem maior oscilação entre os casos.

\begin{table*}[h!]
\centering
\small
\caption{Desempenho médio e desvio padrão dos sistemas de IA em cada um dos critérios de avaliação.}
\rowcolors{2}{lightgreen}{white}
\begin{tabular}{lcccc}
\toprule
\textbf{Critério} & \textbf{Gemini} & \textbf{ChatGPT Gratuito} & \textbf{ChatGPT Plus} & \textbf{Jus IA} \\ 
\midrule
Corretude & $0,60 \overset{+}{-} 0,27$ & $0,66 \overset{+}{-} 0,22$ & $0,66 \overset{+}{-} 0,20$ & $0,84 \overset{+}{-} 0,16$ \\
Completude & $0,54 \overset{+}{-} 0,27$ & $0,56 \overset{+}{-} 0,10$ & $0,57 \overset{+}{-} 0,12$ & $0,63 \overset{+}{-} 0,11$ \\
Confiabilidade & $0,42 \overset{+}{-} 0,25$ & $0,55 \overset{+}{-} 0,18$ & $0,53 \overset{+}{-} 0,16$ & $0,83 \overset{+}{-} 0,18$ \\
Fluidez & $0,66 \overset{+}{-} 0,28$ & $0,64 \overset{+}{-} 0,24$ & $0,63 \overset{+}{-} 0,24$ & $0,77 \overset{+}{-} 0,15$ \\
\bottomrule
\end{tabular}
\rowcolors{2}{}{}
\label{tab:resultado_geral_criterios}
\vspace{-0.5em}
\end{table*}

As diferenças foram mais acentuadas nos critérios de \textit{corretude} e \textit{confiabilidade}, sugerindo que a IA jurídica gerou respostas mais alinhadas ao ordenamento jurídico e com fundamentação mais clara. No critério de confiabilidade, o intervalo entre o melhor e pior desempenho, Jus IA e Gemini respectivamente, chega a quase 98\%, refletindo o impacto que dados curados e de domínio têm no treinamento das IAs. Já o menor intervalo foi observado em completude, com uma diferença de aproximadamente 17\% entre o sistemas com melhor e pior resultado, novamente Jus IA e Gemini. Quando comparados com o Gemini, as duas versões do ChatGPT têm resultados medianos mais consistentes entre os casos analisados. É ainda interessar notar que a perfomance do ChatGPT em suas versões gratuita e paga (Plus) são equivalentes, ressaltando que, para uma aplicação de domínio, nem sempre a versão PRO de uma IA generalista entrega melhores resultados do que a versão disponibilizada grauitamente pelos provedores de IAs.

Nos 15 casos analisados, a performance do Jus IA se destacou especialmente no critério de confiabilidade, em que superou os demais sistemas em 93\% das situações, evidenciando sua capacidade de oferecer respostas bem fundamentadas e juridicamente seguras. Em corretude, também apresentou desempenho robusto, sendo superior em 80\% dos casos. Já em completude, foi a melhor em 60\% das análises, indicando boa cobertura dos elementos relevantes, ainda que com margem para evolução. O critério de fluidez apresentou o resultado mais equilibrado, com Jus IA à frente em 47\% dos casos, sugerindo que a clareza e naturalidade textual foram menos determinantes na comparação geral.

Ao analisarmos as questões 1 e 2 do nosso questionário, ambas voltadas à veracidade das citações, observamos que o Jus IA apresentou 88\% de confiabilidade. Isso significa que, na vasta maioria dos casos, ele recuperou corretamente as referências normativas, sem erros ou alucinações. Na prática, isso se traduz em uma taxa de alucinação 43\% inferior à dos sistemas generalistas avaliados.

Para comprovar que esses resultados fazem diferença na prática, ou ainda, são resultados estáveis e podem ser generalizados, é necessário verificar se o experimento tem relevância estatístisca. A verificação da relevância estatística (ou significância estatística) é uma etapa essencial para confirmar que as diferenças observadas entre os sistemas não ocorreram por acaso. Em outras palavras, ela indica se o desempenho superior de um sistema é consistente o bastante para ser considerado um resultado real, e não uma variação aleatória dos dados. Para isso, utilizam-se testes estatísticos que comparam os resultados de múltiplos sistemas em várias tarefas e verificam se as diferenças nas médias são estatisticamente significativas. Quando há significância, pode-se afirmar, com grau de confiança mensurável, que o desempenho do sistema destacado reflete uma superioridade efetiva e reproduzível, e não apenas circunstancial.

Dessa forma, foram realizados os testes de Friedman~\cite{friedman1937ranks} e Nemenyi~\cite{nemenyi1963distribution}, cujos resultados indicam significância estatística. O teste de Friedman apresentou valor de 28,7, e a análise post-hoc de Nemenyi revelou que o desempenho do sistema Jus IA é estatisticamente superior ao de todos os demais sistemas comparados.



\subsection{Análise por intenção de uso}
\label{sec-intencao}

Após a constatação de superioridade média da IA jurídica nos critérios centrais~\ref{sec-desenpenho-geral}, a análise por intenção de uso detalha como esse desempenho se manifesta em tarefas de natureza diversa. Como apresentado na Seção~\ref{sec-intencao}, as cinco intenções de uso avaliadas foram: Gerar / Ajustar Documento, Analisar / Resumir Documento, Entender Conceito Jurídico, Entender Panorama Jurisprudencial e Pesquisar Precedente.

Tabela~\ref{tab:resultado_geral_intencao} apresenta a média e desvio padrão dos resultados obtidos por cada um dos sistemas nas cinco intenções de uso avaliadas. Como foi apresentado na Seção~\ref{sec:pesos-criterios}, o desempenho de cada sistema para cada caso é calculado seguindo a Fórmula~\ref{eq:desempenho}, onde os critérios \textit{corretude} e \textit{confiabilidade} tem um peso maior que os critérios \textit{completude} e \textit{fluidez}.

\begin{table*}[h!]
\centering
\small
\caption{Desempenho médio e desvio padrão dos sistemas de IA em cada uma das intenções avaliadas.}
\rowcolors{2}{lightgreen}{white}
\begin{tabular}{L{3.5cm}cccc}
\toprule
\textbf{Intenções} & \textbf{Gemini} & \textbf{ChatGPT Gratuito} & \textbf{ChatGPT Plus} & \textbf{Jus IA} \\ 
\midrule
Analisar / Resumir Documentos & $0,24 \overset{+}{-} 0,21$ & $0,28 \overset{+}{-} 0,20$ & $0,25 \overset{+}{-} 0,27$ & $0,80 \overset{+}{-} 0,18$ \\
Entender Conceito Jurídico & $0,30 \overset{+}{-} 0,29$ & $0,44 \overset{+}{-} 0,28$ & $0,41 \overset{+}{-} 0,10$ & $0,89 \overset{+}{-} 0,14$ \\
Entender Panorama Jurisprudencial & $0,12 \overset{+}{-} 0,08$ & $0,24 \overset{+}{-} 0,22$ & $0,40 \overset{+}{-} 0,24$ & $0,83 \overset{+}{-} 0,14$ \\
Gerar / Ajustar Documento & $0,55 \overset{+}{-} 0,45$ & $0,36 \overset{+}{-} 0,34$ & $0,36 \overset{+}{-} 0,30$ & $0,64 \overset{+}{-} 0,43$ \\
Pesquisar Precedentes e Casos Similares & $0,14 \overset{+}{-} 0,14$ & $0,59 \overset{+}{-} 0,27$ & $0,49 \overset{+}{-} 0,12$ & $0,95 \overset{+}{-} 0,09$ \\
\bottomrule
\end{tabular}
\rowcolors{2}{}{}
\label{tab:resultado_geral_intencao}
\vspace{-0.5em}
\end{table*}

Conforme ilustrado na Tabela~\ref{tab:resultado_geral_intencao}, o sistema de IA Jurídico Jus IA obteve as maiores médias nas cinco intenções avaliadas. O destaque vai para a intenção \textit{Pesquisar Precedentes e Casos Similares}, na qual o Jus IA alcançou um desempenho de 0,95, contrastando com o valor de 0,14 registrado pelo sistema Gemini, o mais baixo nessa categoria. Já na intenção \textit{Gerar / Ajustar Documento}, a vantagem do Jus IA em relação aos demais sistemas diminui significativamente. De modo geral, essa foi a intenção que apresentou maior variabilidade de desempenho entre os sistemas analisados.

\paragraph{Analisar / Resumir Documentos} Nesta intenção de uso, os sistemas generalistas demonstraram desempenho competitivo, especialmente nos critérios de \textit{completude} e \textit{fluidez}. O modelo ChatGPT Plus, em particular, apresentou resultados ligeiramente superiores nesses aspectos, evidenciando sua proficiência em tarefas de síntese textual e geração de linguagem natural. A Tabela~\ref{tab:resultado_analisar_resumir} traz as médias e desvio padrão para cada um dos sistemas considerando apenas os casos da intenção de uso \textit{Analisar / Resumir Documentos}.

No entanto, quando avaliados sob os critérios de \textit{corretude} e \textit{confiabilidade}, a solução Jus IA destacou-se com desempenho significativamente superior. Essa diferença sugere que, embora os modelos generalistas sejam eficazes na organização e apresentação fluente de informações, eles tendem a apresentar menor rigor jurídico e menor aderência normativa.

Esse equilíbrio entre completude e fluidez, por um lado, e precisão jurídica, por outro, evidencia a competência dos modelos generalistas em tarefas de abstração e reescrita, ainda que com menor rigor jurídico. Sistemas especializados como Jus IA demonstram maior competência na interpretação e aplicação de conceitos técnicos do Direito. 

\begin{table*}[h!]
\centering
\small
\caption{Desempenho médio e desvio padrão dos sistemas de IA em cada um dos critérios de avaliação considerando a intenção de uso Analisar / Resumir Documentos.}
\rowcolors{2}{lightgreen}{white}
\begin{tabular}{lcccc}
\toprule
\textbf{Critério} & \textbf{Gemini} & \textbf{ChatGPT Gratuito} & \textbf{ChatGPT Plus} & \textbf{Jus IA} \\ 
\midrule
Corretude & $ 0,36 \overset{+}{-} 0,30 $ & $ 0,38 \overset{+}{-} 0,33 $ & $ 0,43 \overset{+}{-} 0,36 $ & $ 0,71 \overset{+}{-} 0,25 $ \\
Completude & $ 0,61 \overset{+}{-} 0,35 $ & $ 0,61 \overset{+}{-} 0,10 $ & $ 0,78 \overset{+}{-} 0,25 $ & $ 0,61 \overset{+}{-} 0,54 $ \\
Confiabilidade & $ 0,23 \overset{+}{-} 0,19 $ & $ 0,27 \overset{+}{-} 0,15 $ & $ 0,20 \overset{+}{-} 0,21 $ & $ 0,87 \overset{+}{-} 0,12 $ \\
Fluidez & $ 0,54 \overset{+}{-} 0,44 $ & $ 0,54 \overset{+}{-} 0,07 $ & $ 0,62 \overset{+}{-} 0,45 $ & $ 0,46 \overset{+}{-} 0,44 $ \\
\bottomrule
\end{tabular}
\rowcolors{2}{}{}
\label{tab:resultado_analisar_resumir}
\vspace{-0.5em}
\end{table*}

\paragraph{Entender Conceito Jurídico} A Tabela~\ref{tab:resultado_conceito_jurídico} apresenta as médias e os respectivos desvios padrão para a intenção de uso \textit{Entender Conceito Jurídico}, considerando os quatro critérios de avaliação. A solução Jus IA obteve os melhores desempenhos em todos os critérios, com destaque expressivo em \textit{confiabilidade} (0,94) e \textit{corretude} (0,77). 

Em contraste, o Gemini apresentou os menores índices em \textit{confiabilidade} e \textit{corretude}, demonstrando menor domínio sobre conceitos técnicos do Direito.

\begin{table*}[h!]
\centering
\small
\caption{Desempenho médio e desvio padrão dos sistemas de IA em cada um dos critérios de avaliação considerando a intenção de uso Entender Conceito Jurídico.}
\rowcolors{2}{lightgreen}{white} 
\begin{tabular}{lcccc}
\toprule
\textbf{Critério} & \textbf{Gemini} & \textbf{ChatGPT Gratuito} & \textbf{ChatGPT Plus} & \textbf{Jus IA} \\ 
\midrule
Corretude & $ 0,29 \overset{+}{-} 0,29 $ & $ 0,45 \overset{+}{-} 0,33 $ & $ 0,35 \overset{+}{-} 0,12 $ & $ 0,77 \overset{+}{-} 0,29 $ \\
Completude & $ 0,61 \overset{+}{-} 0,42 $ & $ 0,50 \overset{+}{-} 0,44 $ & $ 0,72 \overset{+}{-} 0,10 $ & $ 0,89 \overset{+}{-} 0,10 $ \\
Confiabilidade & $ 0,29 \overset{+}{-} 0,28 $ & $ 0,44 \overset{+}{-} 0,26 $ & $ 0,42 \overset{+}{-} 0,08 $ & $ 0,94 \overset{+}{-} 0,09 $ \\
Fluidez & $ 0,52 \overset{+}{-} 0,44 $ & $ 0,43 \overset{+}{-} 0,38 $ & $ 0,43 \overset{+}{-} 0,38 $ & $ 0,71 \overset{+}{-} 0,00 $ \\
\bottomrule
\end{tabular}
\rowcolors{2}{}{}
\label{tab:resultado_conceito_jurídico}
\vspace{-0.5em}
\end{table*}

Entre os comentários registrados pelos avaliadores, a solução Jus IA foi consistentemente elogiada pela solidez argumentativa de suas respostas, especialmente pela incorporação de jurisprudência e dispositivos legais como suporte à descrição conceitual. Um dos advogados chegou a classificá-la como ``tecnicamente excelente'', destacando sua aderência normativa e precisão interpretativa.


Em contraste, observou-se que o Gemini tende a oferecer respostas mais simplificadas, com menor ancoragem em fontes jurídicas primárias, o que pode limitar a profundidade e a confiabilidade das análises em determinados contextos.

\paragraph{Entender Panorama Jurisprudencial} A análise dos resultados apresentados na Tabela~\ref{tab:resultado_panorama_jurisprudencial} revela que o modelo Jus IA se destaca amplamente em tarefas relacionadas à interpretação jurisprudencial, superando os demais em \textit{corretude}, \textit{completude} e \textit{confiabilidade}. Esses critérios são essenciais para garantir precisão normativa, cobertura argumentativa e consistência lógica, pilares fundamentais na análise de decisões judiciais. Além disso, Jus IA apresenta baixa variabilidade nos resultados, o que sugere desempenho estável e auditável.

Os modelos generalistas (ChatGPT Gratuito, ChatGPT Plus e Gemini), embora apresentem algum nível de fluidez, demonstram limitações significativas nos critérios mais críticos para o domínio jurídico. A alta variabilidade em suas métricas indica que, mesmo quando acertam, o desempenho é inconsistente.

O ChatGPT Plus apresentou desempenho superior ao gratuito apenas nesse caso específico de uso; nos demais critérios, a versão gratuita mostrou-se competitiva.


\begin{table*}[h!]
\centering
\small
\caption{Desempenho médio e desvio padrão dos sistemas de IA em cada um dos critérios de avaliação considerando a intenção de uso Entender Panorama Jurisprudencial.}
\rowcolors{2}{lightgreen}{white} 
\begin{tabular}{lcccc}
\toprule
\textbf{Critério} & \textbf{Gemini} & \textbf{ChatGPT Gratuito} & \textbf{ChatGPT Plus} & \textbf{Jus IA} \\ 
\midrule
Corretude & $ 0.25 \overset{+}{-} 0.18 $ &  $ 0.09 \overset{+}{-} 0.16 $ & $ 0.35 \overset{+}{-} 0.28 $ & $ 0.81 \overset{+}{-} 0.17 $ \\
Completude & $ 0.42 \overset{+}{-} 0.43 $ & $ 0.71 \overset{+}{-} 0.21 $ & $ 0.80 \overset{+}{-} 0.12 $ & $ 0.93 \overset{+}{-} 0.07 $ \\
Confiabilidade & $ 0.09 \overset{+}{-} 0.10 $ & $ 0.29 \overset{+}{-} 0.26 $ & $ 0.42 \overset{+}{-} 0.24 $ & $ 0.85 \overset{+}{-} 0.13 $ \\
Fluidez & $ 0.53 \overset{+}{-} 0.50 $ & $ 0.60 \overset{+}{-} 0.00 $ & $ 0.47 \overset{+}{-} 0.50 $ & $ 0.60 \overset{+}{-} 0.53 $ \\
\bottomrule
\end{tabular}
\rowcolors{2}{}{} 
\label{tab:resultado_panorama_jurisprudencial}
\vspace{-0.5em}
\end{table*}

\paragraph{Gerar / Ajustar Documento} Nesta categoria de intenção de uso, observou-se uma convergência relativa nos desempenhos das quatro soluções avaliadas, embora persistam diferenças significativas na qualidade técnica dos conteúdos gerados. A Tabela~\ref{tab:resultado_gerar_ajustar} apresenta as médias e desvio padrão de cada um dos sistemas, considerando os quatro critérios avaliados.

De maneira geral, a solução Jus IA apresentou os melhores resultados em termos de \textit{corretude} e \textit{confiabilidade}, evidenciando maior aderência às normas jurídicas e precisão argumentativa. Por outro lado, o modelo Gemini destacou-se em \textit{fluidez}, produzindo textos mais naturais e linguisticamente refinados. As variantes do ChatGPT exibiram desempenho intermediário, com variações entre \textit{completude} e \textit{fluidez}, porém com maior frequência de imprecisões normativas.

Importa destacar que esta intenção de uso apresentou as menores médias gerais entre os critérios avaliados, o que reforça a necessidade de atuação conjunta entre profissionais jurídicos e sistemas de IA. Tal colaboração é essencial para assegurar a qualidade, precisão e aplicabilidade das soluções propostas.


\begin{table*}[h!]
\centering
\small
\caption{Desempenho médio e desvio padrão dos sistemas de IA em cada um dos critérios de avaliação considerando a intenção de uso Gerar / Ajustar Documento.}
\rowcolors{2}{lightgreen}{white} 
\begin{tabular}{lcccc}
\toprule
\textbf{Critério} & \textbf{Gemini} & \textbf{ChatGPT Gratuito} & \textbf{ChatGPT Plus} & \textbf{Jus IA} \\ 
\midrule
Corretude & $ 0,59 \overset{+}{-} 0,49 $ & $ 0,56 \overset{+}{-} 0,48 $ & $ 0,52 \overset{+}{-} 0,46 $ & $ 0,61 \overset{+}{-} 0,47 $ \\
Completude & $ 0,76 \overset{+}{-} 0,37 $ & $ 0,32 \overset{+}{-} 0,26 $ & $ 0,33 \overset{+}{-} 0,32 $ & $ 0,56 \overset{+}{-} 0,31 $ \\
Confiabilidade & $ 0,32 \overset{+}{-} 0,27 $ & $ 0,20 \overset{+}{-} 0,17 $ & $ 0,23 \overset{+}{-} 0,20 $ & $ 0,52 \overset{+}{-} 0,43 $ \\
Fluidez & $ 0,82 \overset{+}{-} 0,31 $ & $ 0,59 \overset{+}{-} 0,19 $ & $ 0,46 \overset{+}{-} 0,43 $ & $ 0,64 \overset{+}{-} 0,27 $ \\
\bottomrule
\end{tabular}
\rowcolors{2}{}{} 
\label{tab:resultado_gerar_ajustar}
\vspace{-0.5em}
\end{table*}

Na tarefa de elaboração de peças jurídicas, todas as soluções analisadas apresentaram falhas estruturais recorrentes, incluindo inconsistências na qualificação das partes, inadequações no endereçamento e formulação imprecisa dos pedidos.

Quanto à construção argumentativa, o modelo Gemini demonstrou maior capacidade de contextualização dos fatos do caso, embora não tenha incorporado referências jurisprudenciais. Em contraste, a solução Jus IA omitiu elementos factuais relevantes, comprometendo a fundamentação, enquanto o ChatGPT, nas suas duas versões, apresentou sinais de confusão na interpretação dos fatos, resultando em argumentos desconexos ou pouco alinhados ao contexto jurídico.

\paragraph{Pesquisar Precedentes / Casos Similares} Entre os diversos critérios avaliados, a intenção de uso voltada à pesquisa de precedentes e casos similares foi aquela em que o sistema jus IA apresentou desempenho mais expressivo. Com valores próximos da excelência em \textit{corretude} (0.93), \textit{completude} (0.86), \textit{confiabilidade} (0.96) e \textit{fluidez} (0.97), o modelo demonstrou não apenas domínio técnico, mas também consistência argumentativa e clareza textual, atributos essenciais para a recuperação e interpretação de jurisprudência. A Tabela~\ref{tab:resultado_precedentes} apresenta as médias e os respectivos desvios padrão para cada um dos sistemas avaliados. 

Esse resultado é coerente com a arquitetura do Jus IA: um sistema jurídico especializado, treinado sobre a maior base de jurisprudência do Brasil e equipado com uma busca jurídica avançada, capaz de identificar padrões decisórios, extrair fundamentos relevantes e contextualizar casos com precisão. Ao contrário dos modelos generalistas, que tendem a apresentar variabilidade elevada e lacunas normativas, o Jus IA se destaca por sua capacidade de navegar com profundidade e segurança pelo universo jurídico brasileiro.

ChatGPT Gratuito demonstrou completude sólida (0.69), superando o ChatGPT Plus nesse critério. Isso sugere que, mesmo sem acesso premium, o modelo consegue cobrir bem os elementos esperados em uma resposta jurídica. ChatGPT Plus teve desempenho consistente em confiabilidade (0.71), indicando maior estabilidade argumentativa em relação a versão gratuita e ao Gemini.

Gemini teve os piores resultados em todos os critérios, especialmente em \textit{confiabilidade} (0.09) e \textit{completude} (0.21), o que compromete sua utilidade em tarefas jurídicas mais exigentes.

\begin{table*}[h!]
\centering
\small
\caption{Desempenho médio e desvio padrão dos sistemas de IA em cada um dos critérios de avaliação considerando a intenção de uso Pesquisar Precedentes / Casos Similares.}
\rowcolors{2}{lightgreen}{white}
\begin{tabular}{lcccc}
\toprule
\textbf{Critério} & \textbf{Gemini} & \textbf{ChatGPT Gratuito} & \textbf{ChatGPT Plus} & \textbf{Jus IA} \\ 
\midrule
Corretude & $ 0,25 \overset{+}{-} 0,22 $ & $ 0,58 \overset{+}{-} 0,36 $ & $ 0,51 \overset{+}{-} 0,14 $ & $ 0,93 \overset{+}{-} 0,10 $ \\
Completude & $ 0,21 \overset{+}{-} 0,19 $ & $ 0,69 \overset{+}{-} 0,10 $ & $ 0,63 \overset{+}{-} 0,12 $ & $ 0,86 \overset{+}{-} 0,20 $ \\
Confiabilidade & $ 0,09 \overset{+}{-} 0,12 $ & $ 0,60 \overset{+}{-} 0,23 $ & $ 0,47 \overset{+}{-} 0,11 $ & $ 0,96 \overset{+}{-} 0,06 $ \\
Fluidez & $ 0,26 \overset{+}{-} 0,22 $ & $ 0,41 \overset{+}{-} 0,42 $ & $ 0,51 \overset{+}{-} 0,19 $ & $ 0,97 \overset{+}{-} 0,04 $ \\
\bottomrule
\end{tabular}
\rowcolors{2}{}{}
\label{tab:resultado_precedentes}
\vspace{-0.5em}
\end{table*}

Concluindo, os resultados indicam que o sistema Jus IA apresentou desempenho superior em todas as dimensões avaliadas, com destaque para os critérios de \textit{corretude} e \textit{confiabilidade}, onde as diferenças foram mais acentuadas. Embora as soluções generalistas tenham demonstrado competência linguística, especialmente em \textit{fluidez} e \textit{completude}, seu desempenho foi mais instável, com maior variabilidade entre casos e tendência a incorreções jurídicas.

Esses resultados são reforçados em justificativas apresentadas pelos avaliadores, que evidenciam três padrões principais:
\begin{enumerate}
    \item Fundamentação explícita: As respostas da IA jurídica consistentemente citaram as normas corretas e explicaram o raciocínio de forma estruturada.
    \item Erros conceituais: Os modelos generalistas confundiram conceitos jurídicos relacionados, como por exemplo \textit{prescrição} e \textit{decadência}.
    \item Estilo comunicativo: O texto das ferramentas generalistas é mais fluido, mas menos preciso juridicamente.
\end{enumerate}


\subsection{Interpretação dos resultados à luz da prática jurídica}

Os resultados empíricos confirmam a hipótese inicial: a especialização jurídica melhora de modo consistente a correção e a confiabilidade das respostas, mantendo níveis de fluidez equivalentes aos dos modelos generalistas. Isso sugere que a competência linguística é hoje um ponto de convergência entre sistemas, enquanto a qualidade jurídica depende de fatores mais estruturais, como precisão normativa, coerência argumentativa e rastreabilidade das fontes.

Do ponto de vista epistemológico, o estudo demonstra que a avaliação jurídica não pode ser reduzida a métricas automáticas. O julgamento humano, orientado por princípios jurídicos, continua sendo insubstituível para aferir qualidade normativa e argumentativa~\cite{gao2024}.

Metodologicamente, o experimento valida a viabilidade de protocolos de avaliação empírica guiados por fundamentos do Direito, oferecendo um modelo replicável para outras jurisdições.

\section{Conclusão e Trabalhos Futuros}

Este estudo apresentou e aplicou o protocolo do \textit{Índice de Qualidade Jus IA}, uma proposta metodológica para a avaliação de sistemas de Inteligência Artificial (IA) no domínio jurídico.
O experimento comparou quatro sistemas de IA conversacional baseados em LLMs: uma solução jurídica especializada, o Jus IA, e três sistemas de propósito geral, ChatGPT Gratuito, ChatGPT Plus e Gemini.

O protocolo do \textit{Índice de Qualidade Jus IA} mede a qualidade de respostas jurídicas segundo quatro critérios técnicos estabelecidos em conformidade às dimensões jurídico-argumentativas mais relevantes para o dia-a-dia do profissional de Direito. O protocolo foi desenvolvido de maneira multisciplinar, considerando expertise jurídica e expertise em Inteligência Artificial e protocolos de avaliação.

Foi realizado um experimento avaliando as respostas das quatro IAs em 15 cenários reais. O conjunto de análise final contém 60 respostas, que foram analisadas cegamente por 48 profissionais de Direito, com backgrounds diversos. Em linhas gerais, o experimento mostra que a especialização das Inteligências Artificiais para o domínio do Direito melhora consideravelmente a performance dos sistemas. A Inteligência especializada, Jus IA, reduziu em 43\% as alucinações dos modelos. A IA Especializada também foi a melhor colocada considerando todos os critérios juridicamente motivados, evidenciando o papel indispensável da expertise jurídica no desenvolvimento de tais sistemas.

Os resultados reforçam que a especialização é determinante para a qualidade jurídica: modelos treinados com base normativa e dogmática, em dados curados por especialistas, produzem respostas mais corretas e confiáveis. 

As contribuições deste trabalho são metodológicas, conceituais e práticas:

\begin{itemize}
    \item metodológica, ao propor um protocolo de avaliação replicável, que considera a estrutura do pensamento jurídico para a definição e valoração dos critérios de avaliação;
    \item conceitual, ao demonstrar que a racionalidade jurídica pode ser considerada na construção de critérios de qualidade para sistemas de IA especializados no domínio;
    \item pragmática, ao demonstrar que, no momento atual da tecnologia, IAs especializadas superam enormemente IAs generalistas no domínio jurídico em português do Brasil.
\end{itemize}

Trabalhos futuros incluem a ampliação do conjunto de casos, bem como a integração de métricas automáticas que permitam escalar o experimento de forma mais robusta. Também está prevista a replicação da metodologia com outras ferramentas especializadas, além do Jus IA, incluindo LLMs open-source e modelos aplicados a diferentes jurisdições. Por fim, propõe-se a evolução do \textit{Índice de qualidade Jus IA} como um benchmark público e iterativo, contribuindo para práticas de governança responsável na aplicação da IA jurídica.

Pretende-se com esse trabalho não apenas comparar diferentes IAs em cenários reais, mas principalmente estabelecer um protocolo de avaliação aberto, sistemático e reproduzível para sistemas jurídicos baseados em inteligência artificial. Esperamos que o \textit{Índice Jus IA} possa ser reutilizado em pesquisas futuras, acadêmicas ou industriais, de forma a padronizar as análises deste domínio tão crítico rumo à democratização e à integridade do uso da Inteligência Artificial no Direito. Ao oferecer uma metodologia aberta, replicável e ancorada em critérios jurídicos, o estudo busca contribuir para práticas mais transparentes de avaliação e para o fortalecimento de uma governança pública da IA jurídica no Brasil.

\section{Limitações}
Apesar dos resultados expressivos e da validação metodológica do protocolo Jus IA, o estudo apresenta algumas limitações que devem ser reconhecidas. Em primeiro lugar, o número de casos e de intenções de uso avaliados foi limitado a quinze situações, o que restringe a generalização dos achados a outros tipos de tarefa jurídica. Além disso, o experimento concentrou-se exclusivamente no contexto brasileiro, utilizando corpora normativos e doutrinários nacionais. Novos experimentos são necessários para confirmar a aplicabilidade do protocolo Jus IA em outras jurisdições ou sistemas jurídicos de tradição distinta.

Este estudo foi realizado em setembro de 2025; portanto, evoluções de cada um dos sistemas de IA depois desta data não poderiam ser consideradas.

Outro ponto relevante diz respeito à dependência da avaliação humana. Embora o experimento tenha levado em consideração estratégias para a mitigação da subjetividade da avaliação humana, como o cegamento duplo e a randomização da ordem dos textos avaliados, a interpretação dos avaliadores continua sujeita a variações subjetivas próprias da prática jurídica.

Por fim, esse estudo limitou-se a investigar a performance de LLMs proprietárias e de larga escala considerando o propósito maior do trabalho: avaliar os modelos estado da arte generalistas em comparação a um modelo robusto de domínio. A avaliação da performance de modelos pequenos (small models) ou modelos abertos para as mesmas tarefas permanece um trabalho futuro.

\section{Declaração Ética}

Este trabalho seguiu princípios éticos em todas as etapas de coleta, tratamento e análise dos dados. As interações utilizadas foram previamente tratadas em conformidade com a Lei Geral de Proteção de Dados (Lei nº 13.709/2018) e integralmente anonimizadas, de modo que nenhuma informação pessoal ou identificável estivesse disponível aos sistemas avaliados ou às pessoas avaliadoras.

Os 48 profissionais participantes foram convidados individualmente, informados sobre o objetivo geral da pesquisa (avaliar respostas para perguntas de cunho jurídico), mas não foram instruídos sobre a geração automática das respostas, nem tiveram acesso a quais modelos geraram quais respostas. Os profissionais aceitaram participar dessa pesquisa de forma voluntária e confidencial. Para reconhecer o tempo e a contribuição técnica despendida, os avaliadores foram devidamente remunerados.

Este estudo foi realizado por profissionais ligados à Jusbrasil. Para assegurar a idoneidade da pesquisa, o protocolo experimental seguiu boas práticas de pesquisa envolvendo julgamento humano, incluindo cegamento duplo, randomização das respostas, padronização das instruções e avaliação das respostas exatamente como foram geradas por cada um dos sistemas, assegurando neutralidade e equidade no processo avaliativo.

A presente pesquisa foi intencionalmente realizada de forma transdisciplinar, conduzida por profissionais com background em Direito, Ciências da Computação, Negócios e Linguística. 

\section{Agradecimentos}

Gostaríamos de expressar nossa sincera gratidão a Ana Beatriz Machado, Ana Rita Nascimento, Bruna Bueno, Carolina Santos, Clévia Cristina, Luana Limas, Ludmila Dutra, Luiz Azevedo, Luiza Galuppo, Marcus Nascimento, Mohara Coimbra, Rafael Galassi, Raíra Cavalcanti, Pedro H. Oliveira, bem como aos diversos times e áreas do Jusbrasil pela valiosa contribuição ao desenvolvimento deste estudo. Seu envolvimento foi essencial para a estruturação do dataset, a definição dos critérios de avaliação, a revisão metodológica, e seu comprometimento com a qualidade e a colaboração teve impacto direto nos resultados aqui apresentados.


\appendix
\section*{Apêndice A – Questionário de Avaliação}
\label{sec:apendiceA}

Este apêndice apresenta o conjunto de questões utilizadas no formulário de avaliação aplicado ao estudo descrito neste artigo. As perguntas foram elaboradas com o objetivo de verificar os critérios \textit{corretude}, \textit{completude}, \textit{confiabilidade} e \textit{fluidez}. Cada item busca capturar aspectos específicos da fundamentação jurídica, da contextualização temporal dos precedentes e da transparência argumentativa.

O formulário foi estruturado para permitir respostas categóricas (SIM, NÃO, N/A), acompanhadas de justificativas quando necessário (campo ``Observações''), facilitando tanto a replicabilidade do processo quanto a análise estatística dos resultados. A seguir, são listadas as questões que compõem o instrumento de avaliação.

\begin{itemize}
    \item Correção jurídica e fática
        \begin{enumerate}
            \item  \textbf{Sem erro jurídico ou fático?} Avalia se a resposta preserva a veracidade dos fatos e a correção dos argumentos jurídicos utilizados. Considera-se que todas as proposições estão corretas (ex.: sem interpretações ou paráfrases equivocadas, sem confundir tese do tribunal com argumento de parte e sem trocar instâncias).
                \begin{itemize}
                    \item[] SIM = todas as proposições estão corretas.
                    \item[] NÃO = existe pelo menos uma proposição errada.
                \end{itemize}
            \item \textbf{As referências normativas e jurisprudenciais apresentadas são todas verdadeiras (sem alucinação)?} Todas as leis, artigos, súmulas, acórdãos ou regras processuais citadas existem de fato e foram usadas corretamente, sem invenção ou distorção do seu sentido original.
                \begin{itemize}
                    \item[] SIM = todas as referências são verdadeiras, ou seja, não houve alucinação.
                    \item[] NÃO = houve pelo menos uma alucinação de norma ou jurisprudência.
                    \item[] N/A = não há citação de norma ou jurisprudência.
                \end{itemize}
            \item \textbf{A jurisdição e a hierarquia estão corretas?} As normas e precedentes usados são da jurisdição solicitada e respeitam a hierarquia (tribunais superiores acima dos locais).
                \begin{itemize}
                    \item[] SIM = estão corretos.
                    \item[] NÃO = não respeitam a jurisdição solicitada ou a hierarquia.
                    \item[] N/A = não foram citadas normas ou precedentes.
                \end{itemize}
        \end{enumerate}
    \item Completude (estrutura e aderência)
        \begin{enumerate}[resume]
            \item \textbf{A resposta atendeu às instruções do usuário?} Avalia se o conteúdo cobre tema, escopo, perguntas específicas, restrições e formato solicitado; ainda que a resposta final seja contrária ao resultado desejado pelo usuário (ex.: usuário pedir, por exemplo, jurisprudência que não existe e a resposta ser: não encontramos decisões nesse sentido).
                \begin{itemize}
                    \item[] SIM = todas as instruções e/ou perguntas foram atendidas.
                    \item[] NÃO = existe pelo menos uma instrução ou pergunta ignorada.
                \end{itemize}
            \item \textbf{A peça tem estrutura completa?} Inclui: cabeçalho/qualificação (partes, juízo), fatos, fundamentação, pedidos e fechamento (local/data/assinatura, quando aplicável).
                \begin{itemize}
                    \item[] SIM = a estrutura está completa.
                    \item[] NÃO = falta pelo menos uma parte obrigatória da estrutura.
                    \item[] N/A = não se trata de peça.
                \end{itemize}
        \end{enumerate}
        
    \item Confiabilidade
        \begin{enumerate}[resume]
            \item \textbf{Cada afirmação jurídica trouxe referência à fonte verificável?} Toda proposição jurídica tem citação imediata que a sustente.
                \begin{itemize}
                    \item SIM = todas as afirmações têm fonte (lei, precedente ou doutrina) verificável.
                    \item NÃO = há pelo menos uma afirmação sem referência ou que a fonte não foi encontrada.
                \end{itemize}
            \item \textbf{As fontes usadas são primárias/oficiais?} Preferência por lei, decisão ou súmula oficial. Evitar fontes secundárias: artigos, notícias ou citações sem link que não podem ser verificadas.
                \begin{itemize}
                    \item SIM = as fontes são primárias e/ou oficiais.
                    \item NÃO = há pelo menos uma fonte secundária.
                    \item N/A = não foram referenciadas as fontes usadas.
                \end{itemize}
            \item \textbf{A citação sustenta exatamente o que foi afirmado?} O trecho citado realmente confirma a proposição (sem distorções, sem precedente contrário ou superado).
                \begin{itemize}
                    \item SIM = todas as citações sustentam o afirmado.
                    \item NÃO = há pelo menos uma citação que não sustenta o que foi afirmado.
                    \item N/A = não há citações.
                \end{itemize}
            \item \textbf{Há jurisprudência recente (últimos 5 anos)?} Pelo menos 1 precedente relevante $\leq$ 5 anos ou justificativa válida se apenas precedente mais antigo.
                \begin{itemize}
                    \item[] SIM = há precedente recente ou justificativa válida.
                    \item[] NÃO = só há precedentes antigos sem justificativa.
                    \item[] N/A = não há citação de precedentes ou a citação não é verificável.
                \end{itemize}
        \end{enumerate}
        
    \item Fluidez da linguagem (clareza/forma)
        \begin{enumerate}[resume]
            \item \textbf{Precisão e concisão} O texto transmite a ideia de forma simples e direta, sem redundâncias, coloquialismos, adjetivos desnecessários ou termos vagos (“parece que”, “talvez”) quando existe regra clara.
                \begin{itemize}
                    \item[] SIM = é integralmente claro, conciso e bem estruturado.
                    \item[] NÃO = tem pelo menos um trecho confuso, repetitivo ou impreciso.
                \end{itemize}
        \end{enumerate}
    \item Observações
\end{itemize}



\end{document}